\RequirePackage{amsmath}         % Amsmath package for maths and equations

\documentclass[runningheads,a4paper]{llncs}

\usepackage{graphicx}            % For pdf/bitmapped graphics files
\usepackage{amssymb}             % Maths symbols
\usepackage{amsfonts}            % To get the \mathbb alphabet
\usepackage{enumitem}            % For customisation of enumerate environments
\usepackage{multirow}            % For multi-row cells in tables
\usepackage{siunitx}             % For SI units
\usepackage{url}                 % For URL typesetting and line breaking
\usepackage{xspace}              % For smart spaces in commands
\usepackage[T1]{fontenc}         % For special characters such as '_'
\usepackage{hyperref}            % For hyperlinks
\usepackage{wrapfig}
\usepackage{todonotes}
\usepackage{units}
\usepackage[firstpage=true]{background}

\graphicspath{{images/}}
\DeclareGraphicsExtensions{.pdf,.png,.jpg,.jpeg}

\newcommand{\seclabel}[1]{\label{sec:#1}}

\newcommand{\figlabel}[1]{\label{fig:#1}}

\newcommand{\figref}[1]{Fig.~\ref{fig:#1}\xspace}

\newcommand{\nop}{NimbRo\protect\nobreakdash-OP\xspace}

\newcommand{\cm}{CM730\xspace}
\newcommand{\cmnew}{CM740\xspace}

\newcommand{\iguhop}{igus\textsuperscript{\tiny\circledR}$\!$ Humanoid Open Platform\xspace}

\newcommand\copyrighttext{%
	\parbox{\textwidth}{
		\footnotesize
		\textbf{Accepted final version.} In \textit{Proceedings of 21th RoboCup International Symposium, Nagoya, Japan.}
		DOI: \href{https://www.springer.com/us/book/9783030003074}{No. 10.1007/978-3-030-00308-1}
	}
}

\SetBgContents{\copyrighttext}
\SetBgScale{1}
\SetBgColor{black}
\SetBgAngle{0}
\SetBgOpacity{1}
\SetBgPosition{current page.south}
\SetBgVshift{4cm}
\SetBgHshift{3mm}

\begin{document}

% Start of the main matter
\mainmatter

% Paper title
\title{Advanced Soccer Skills and Team Play of RoboCup 2017 TeenSize Winner NimbRo}
\titlerunning{RoboCup 2017 TeenSize Winner NimbRo}

% Authors
\author{Diego Rodriguez, Hafez Farazi, Philipp Allgeuer, Dmytro Pavlichenko, Grzegorz Ficht, Andr\'{e} Brandenburger, Johannes K\"{u}rsch and Sven Behnke}
\authorrunning{Rodriguez, Farazi, Allgeuer, et.al.}

% Institutes
\institute{Autonomous Intelligent Systems, Computer Science, Univ.\ of Bonn, Germany\\
\url{{rodriguez, farazi, pallgeuer, pavlichenko, ficht, behnke}@ais.uni-bonn.de},
\url{http://ais.uni-bonn.de}}

% Typeset the paper title
\maketitle
%%%%%%%%%%%%%%%%%%%%%%%%%%%%%%%%%%%%%%%%%%%%
\begin{abstract}
In order to pursue the vision of the RoboCup Humanoid League of beating the soccer world champion by 2050, new rules and competitions are added or modified each year fostering novel technological advances.
In 2017, the number of players in the TeenSize class soccer games was increase to 3 vs. 3, which allowed for more team play strategies. 
Improvements in individual skills were also demanded through a set of technical challenges.
This paper presents the latest individual skills and team play developments used in RoboCup 2017 that lead our team Nimbro winning the 2017 TeenSize soccer tournament, the technical challenges, and the drop-in games.
\end{abstract}

%%%%%%%%%%%%%%%%%%%%%%%%%%%%%%%%%%%%%%%%%%%%
\section{Introduction}

%-----------------------------------
\begin{figure}[!b]
\centering
\includegraphics[height=45mm]{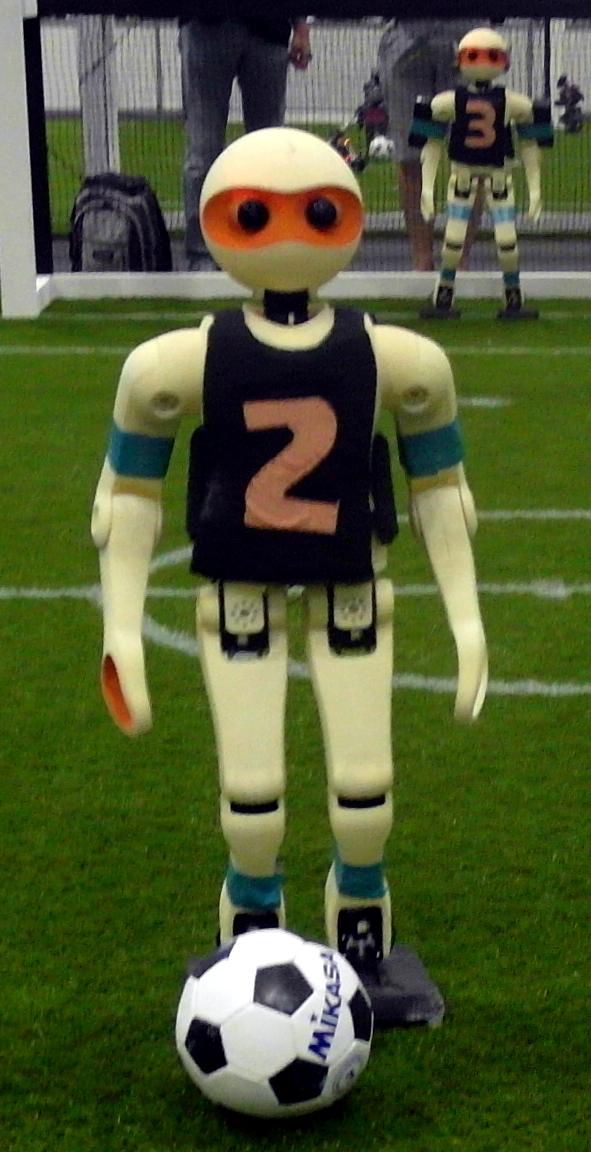}\hspace{1.3mm}%
\includegraphics[height=45mm]{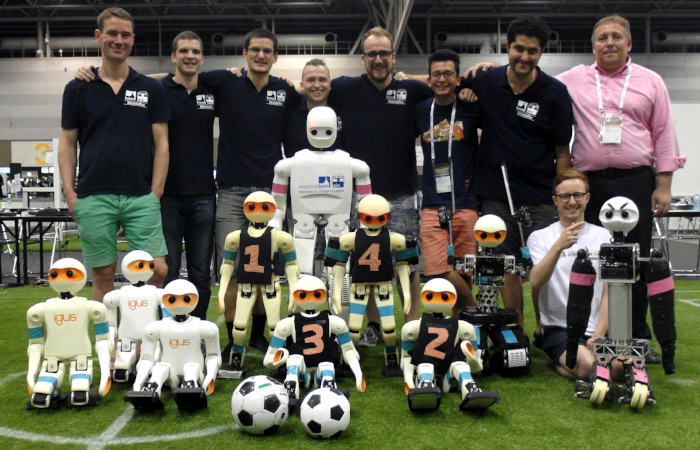}\hspace{1.3mm}%
\includegraphics[height=45mm]{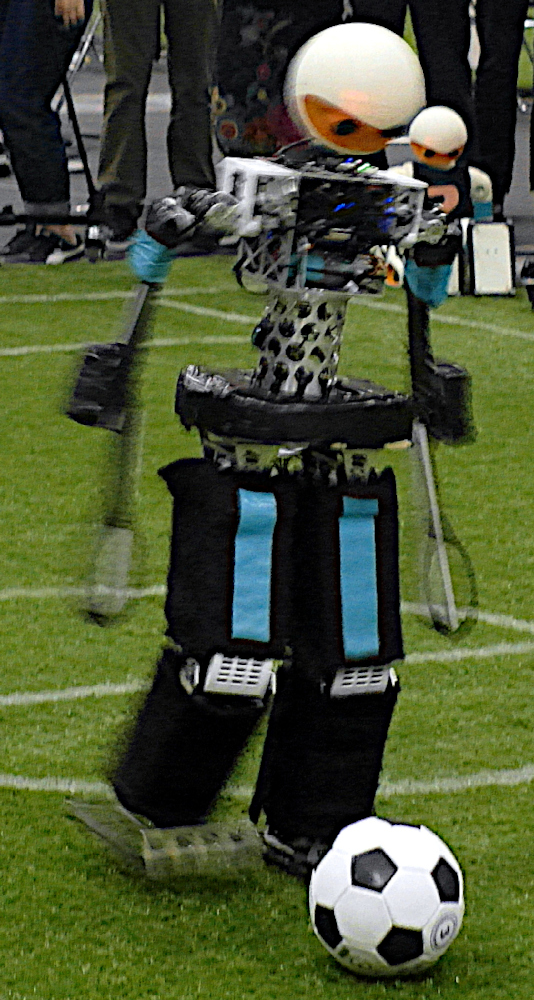}
\caption{Left: the \iguhop robot. Middle: the team NimbRo. Right: the upgraded Dynaped robot for RoboCup 2017.}
\figlabel{nimbro_team}
\vspace{-2ex}
\end{figure}
%-----------------------------------

Every year the RoboCup Humanoid League raises its bar in its competitions.
This year, the league increased the allowed number of players in the TeenSize soccer games to 3 vs. 3, which encourages the development of more complex team play strategies.
At the same time, a new competition called \textit{Drop-in games} was introduced in which a team is composed of robots from different institutes or universities.
This paper presents our recent developments to address these modifications and shows their performance in the competition.
Our robots won all TeenSize 2017 competitions: the soccer tournament, the Drop-in games and the technical challenges.
In RoboCup 2017 we used our fully open-source 3D printed platform, the \iguhop \cite{Allgeuer2015b}.
Moreover, we upgraded one of our classic robots, Dynaped, so that it is able to get up from the prone and supine lying positions in order to be rule-compliant. 
Both platforms are shown in \figref{nimbro_team}, along with the human members of our team NimbRo.
We released a video of the competition 2017 highlights \footnote{RoboCup 2017 NimbRo TeenSize highlights video: \url{https://youtu.be/6ldHWWHfeBc}}.

%%%%%%%%%%%%%%%%%%%%%%%%%%%%%%%%%%%%%%%%%%%%
\section{Robot Platforms}
\seclabel{robot_platforms}
%===============================================================================
\subsubsection{Igus Humanoid Open Platform}
\seclabel{igus}
Over the last four years, the \iguhop \cite{Allgeuer2015b} (\figref{nimbro_team} left) has been developed as an open-source hardware and software project.
Thanks to its 3D printed exoskeleton, the \unit[92]{cm} tall robot weights only \SI{6.6}{\kilogram}.
The platform incorporates an Intel Core i7-5500U CPU running a 64-bit Ubuntu OS, and a Robotis \cm microcontroller board, which electrically interfaces with its Robotis Dynamixel MX actuators. 
The \cm also incorporates 3-axis accelerometer and gyroscope sensors, for a total of 6 axes of inertial measurement. 
For visual perception, the robot is equipped with a Logitech C905 USB camera fitted with a wide-angle lens. 
The robot software is based on the ROS middleware, and is a continuous evolution of the ROS software that was written for the \nop \cite{Allgeuer2013a}.
%===============================================================================
\subsubsection{Dynaped}
\seclabel{dynaped}
Dynaped has been playing since RoboCup 2009 for team NimbRo. 
Its original design had 14 degrees of freedom (DOF): 5\,DoF per leg, 1\,DoF per arm, and 2\,DoF in the neck. 
Dynaped is distinguished by its effective use of parallel kinematics coupled with high torque, provided by pairs of EX-106 actuators in the roll joints of the hip and ankle, and pitch joints in the knee. 
All other DoFs are driven by single actuators. 
The torso is constructed entirely of aluminum and consists of a cylindrical tube and a rectangular cage that holds the electronics.
Similar to the \iguhop, Dynaped is equipped with an Intel Core i7-5500U CPU, a \cmnew controller board (newer version of \cm), and a USB camera with a wide-angle lens.
Dynaped used the same software components as the \iguhop such as perception, bipedal walking, team communication, soccer behaviors, among others, thanks to their modularity and robustness.
Dynaped's competition performance and hardware design, contributed to NimbRo winning the Louis Vuitton Best Humanoid Award in both 2010 and 2012~\cite{Missura:RoboCup2012}.

In order to be allowed to play in RoboCup 2017, Dynaped needed to be upgraded.
Having only 1\,DOF per arm, Dynaped was not able to get up either from prone or supine lying position.
We included thus an additional DOF in each arm, namely a pitch elbow.
Both the arm and the forearm are made from carbon fiber and reinforced against torsion with aluminum bars (\figref{nimbro_team} right).

%%%%%%%%%%%%%%%%%%%%%%%%%%%%%%%%%%%%%%%%%%%%
\section{Software Design}
\seclabel{software_design}

%===============================================================================
\subsection{Visual Perception}
\seclabel{perception}
As part of the preprocessing, we convert the RGB images to the HSV space because of its intuitive nature and handful separation of brightness and chromatic information. 
To compensate the high distortion coming from the wide-angle lenses, we use the pinhole camera model to compensate radial and tangential distortion.
Using this distortion model, once the object of interest is identified, we project it into egocentric world coordinates.
For calibrating the pose of the camera, we use the Nelder-Mead method \cite{nelder1965simplex} to minimize the reprojection error of detected field lines. 
%-------------------------------------------------------------------------------
\vspace{-2ex}
\subsubsection{Ball detection}
We utilize a histogram of oriented gradients (HOG) descriptor in a cascade classifier trained using AdaBoost with positive and negative samples.
Because of the computational cost of using sliding windows \cite{dalal2006object} we use the descriptor only on those candidates that are preselected based on shape, size, and colour.
%-------------------------------------------------------------------------------
\vspace{-2ex}
\subsubsection{Line detection}
\begin{figure}[b]
\centering
\includegraphics[width=0.345\linewidth]{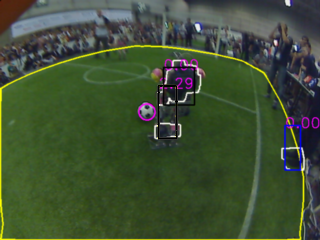} %\hspace{0.019\linewidth}
\includegraphics[width=0.365\linewidth]{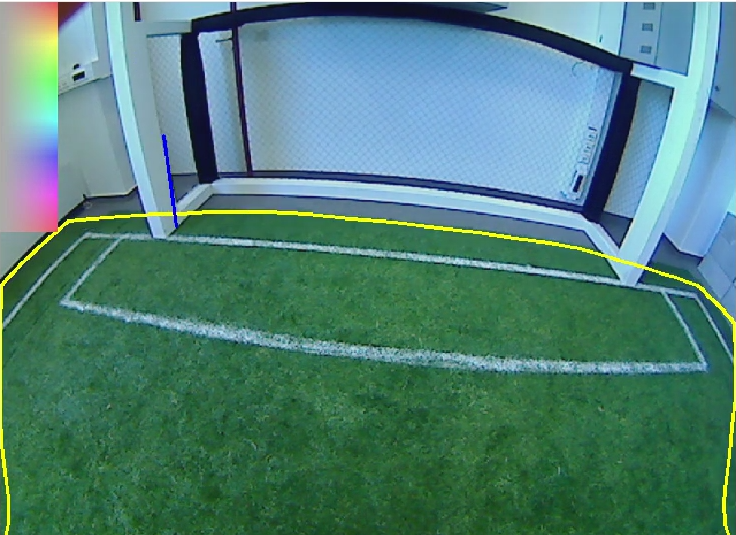}
\includegraphics[width=0.265\linewidth]{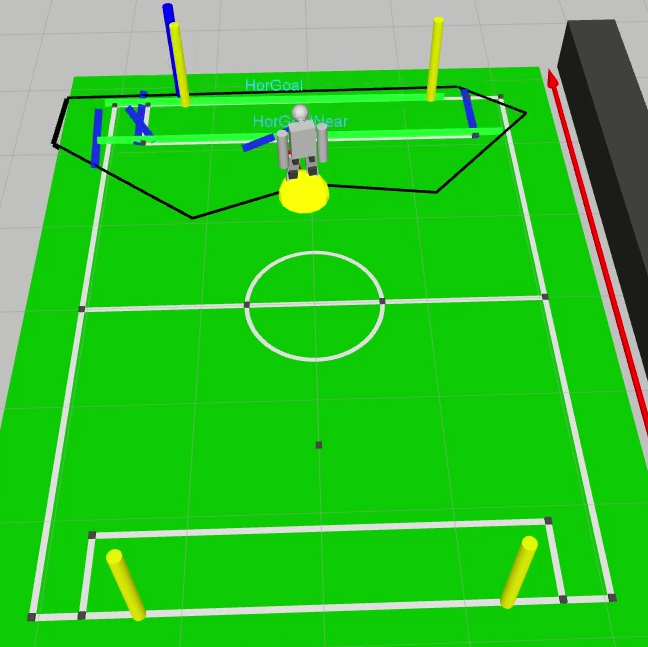}
\caption{Left: Image with ball (magenta circle), obstacles (black square) and field boundary (yellow lines) detections on RoboCup 2017.
Middle: One input image for our localization method.
Right: Visualization of the localization of the robot in Rviz.}
\label{fig:vision_and_loc}
\vspace{-2ex}
\end{figure}

On artificial grass, the painted field lines are not clearly white; thus an edge detector followed by probabilistic Hough line detection are implemented.
Line segments are filtered to avoid false positives.
Finally, the remaining similar line segments are merged to produce fewer larger lines.
The shorter line segments are used for detecting the field circle, while the remaining lines are passed to the localization method.
Figure ~\ref{fig:vision_and_loc} shows an example of the output of the object detection.
For more details, please refer to \cite{Farazi2015}.

%-----------------------------------
%-----------------------------------

%-------------------------------------------------------------------------------
\subsubsection{Localization on the Soccer Field:}
We propose a multi-hypothesis model to estimate the three-dimensional robot pose $(x, y, \theta)$ on the field. 
We mainly use integrated gyroscope values as the source of orientation information 
and treat the heading initialization a classification problem making use of the specific possible locations where the robot can be placed according to the rules, namely, at the touch-line in the robot's own half facing the field, or near the center circle and goal area facing the opponent goal.
We start four instances of our localization module with different initial hypothesis locations.
For each instance, we handle the unknown data association of ambiguous landmarks, such as goal posts and T-junctions.
Over time, we try to update the location of the hypothesis towards an estimated position.
We update the location based on a probabilistic model of landmark observations involving mainly lines and goal posts \cite{Farazi2015}. 
The inputs to the probabilistic model come from the vision module and dead-reckoning odometry data. 
Finally, the hypothesis with the highest probability prevails and the others are deleted.
A sample output of the localization is shown in Fig. \ref{fig:vision_and_loc}.

%-------------------------------------------------------------------------------
\subsection{Bipedal Walking}
The gait of our robots is formulated in three different spaces: joint space, inverse (Cartesian) space and abstract space.
The last one is a convenient formulation for humanoid robots for balancing and walking presented in \cite{Behnke2006}.
The central pattern generator based gait is an extension of our previous work \cite{Missura2013a}.
Starting from a halt pose defined in the abstract space, the central pattern adds waveforms features like leg lifting, leg swinging, arm swinging, among others.
The resulting abstract configuration is then transformed to the inverse space where more motion components are added.
The result is finally converted into joint space to command the actuators.
%-----------------------------------
\begin{figure*}[b]
\vspace{-2ex}
\parbox{\linewidth}{
\centering
\includegraphics[height=24.7mm]{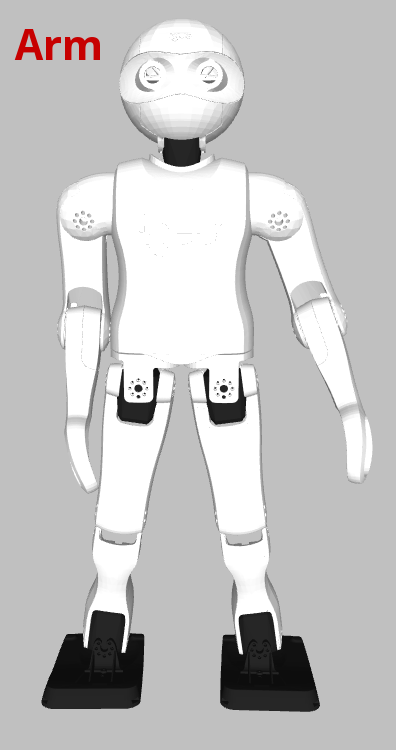}%
\includegraphics[height=24.7mm]{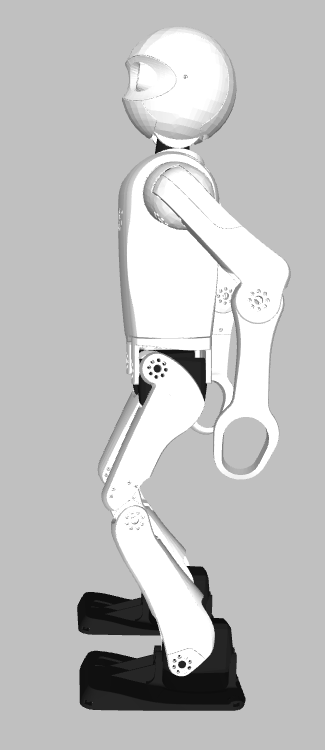}\hspace{2pt}%
\includegraphics[height=24.7mm]{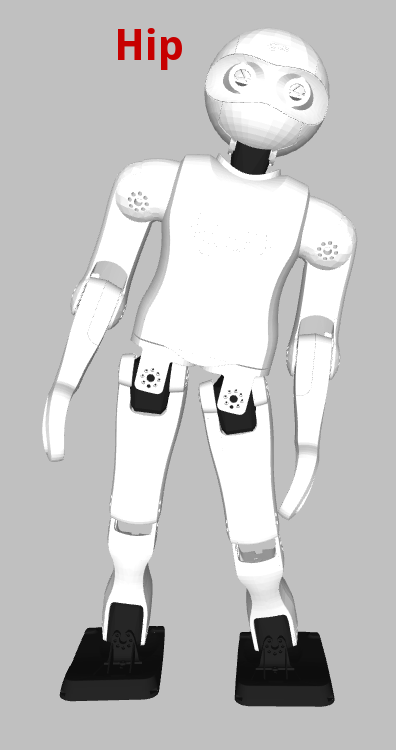}%
\includegraphics[height=24.7mm]{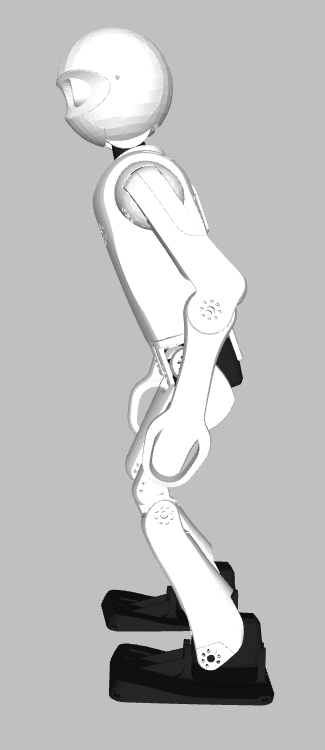}\hspace{2pt}%
\includegraphics[height=24.7mm]{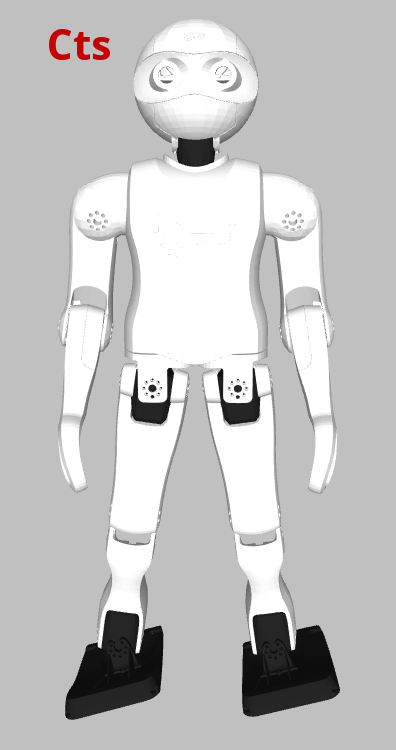}%
\includegraphics[height=24.7mm]{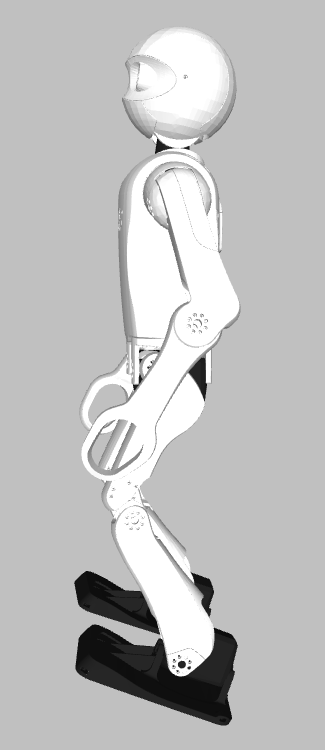}\hspace{2pt}%
\includegraphics[height=24.7mm]{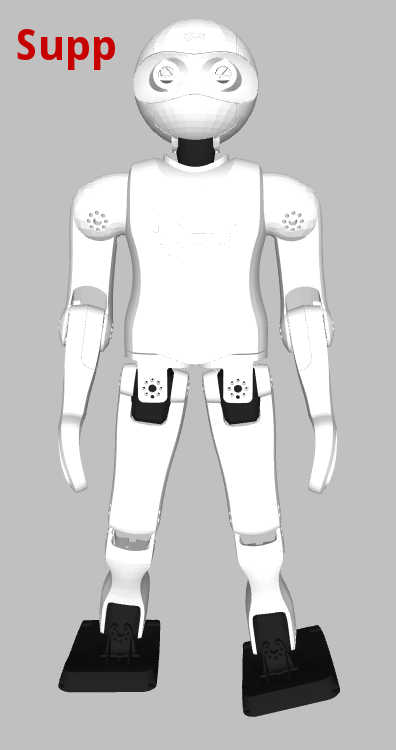}%
\includegraphics[height=24.7mm]{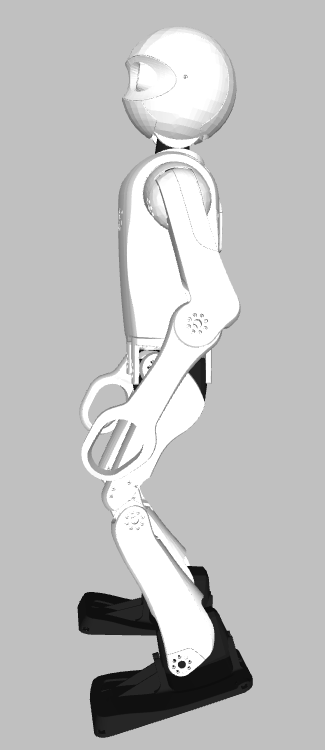}\hspace{2pt}%
\includegraphics[height=24.7mm]{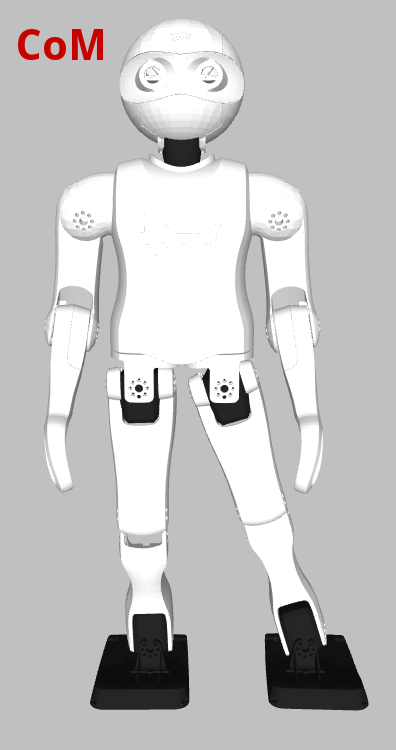}%
\includegraphics[height=24.7mm]{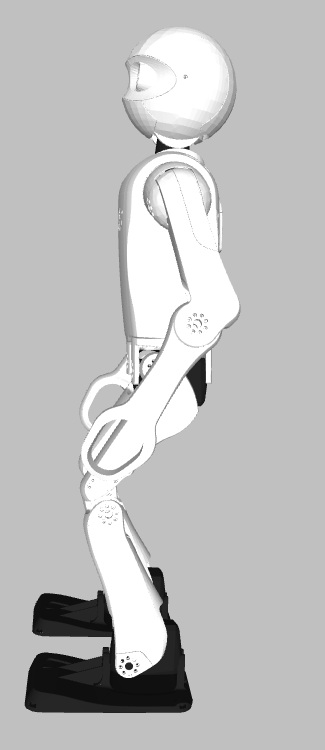}
}
\caption{Corrective actions in the sagittal and lateral planes, from left to right: arm angle, hip angle, continuous foot angle, support foot angle, and CoM shifting actions. 
The actions have been exaggerated for clearer illustration.}
\label{fig:corrective_actions}
\vspace{-4ex}
\end{figure*}
%-----------------------------------
Several feedback mechanisms have been implemented on top of the open-loop gait that help to stabilize the robot \cite{Allgeuer2016a}.
Each of these mechanisms acts as a PID-feedback controlling the fused angle deviations of the robot by adding corrective actions to the central pattern generated waveforms.
The fused angles are an intuitive representation of orientations that offer benefits over Euler angles for balance \cite{Allgeuer2015}.
These mechanisms, illustrated in Fig. \ref{fig:corrective_actions}, include arm angle, hip angle, continuous foot angle, support foot angle and CoM shifting.
The step timing is computed using our capture step framework \cite{missura2014balanced}, based on the lateral CoM state \cite{Allgeuer2016a}.

%-------------------------------------------------------------------------------
\subsection{Soccer Behaviors}

Based on the visual perception of the game state, including ball detections, 
obstacle detections, and the estimated pose of the robot on the field, our robots 
must decide on and execute a strategy for playing soccer. 
This primarily involves localizing the ball and scoring a goal while avoiding obstacles, but 
also extends to team communications, team play, and coordination of the game 
using the information from the RoboCup Humanoid league game controller. A custom 
two-layered hierarchical finite state machine (FSM) has been implemented for 
this purpose and runs in a separate behaviors node. The lower of these two 
layers is referred to as the Behavior FSM, and is responsible for implementing 
low-level skills such as searching for the ball, going to the ball, dribbling 
and kicking. The upper layer is referred to as the Game FSM, and builds on the 
skills implemented in the lower layer to implement game behaviors such as 
default ball handling, which attempts to kick or dribble a ball into goals, and 
positioning, which is used for the auto-positioning setup phase during a 
kickoff. In general, the game states combine groups or sequences of skills to 
execute certain soccer game state-specific behaviors.

%%%%%%%%%%%%%%%%%%%%%%%%%%%%%%%%%%%%%%%%%%%%
\subsection{Team Play}
\begin{figure}[t]
	\centering
	\includegraphics[width=0.65\linewidth]{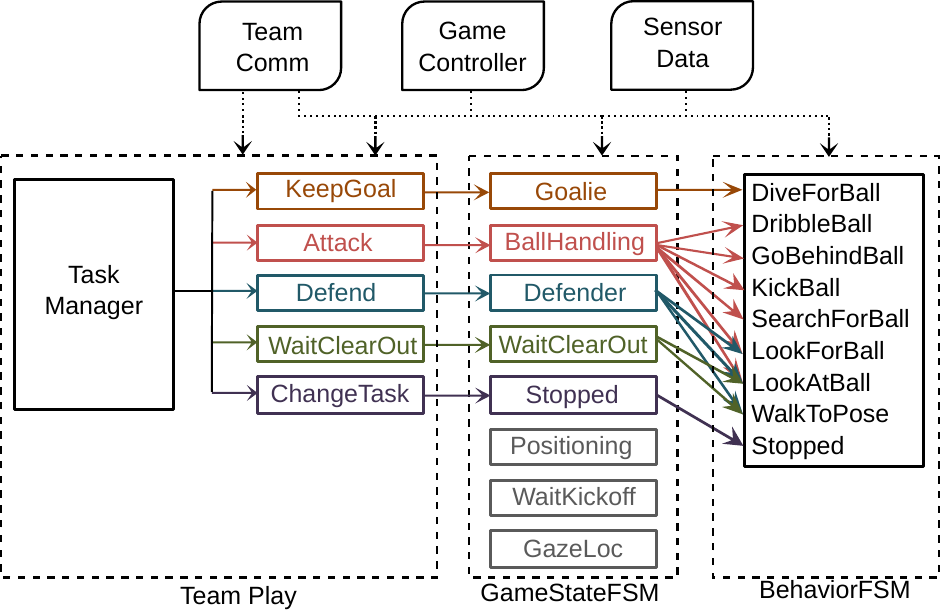} 
	\caption{Based on sensor data, the game controller and the team communication, the task manager assigns a task to each robot. Each task is associated to a state of the Game FSM, which triggers a sequence of behaviors in a lower-level Behavior FSM.}
	\label{fig:roles_tasks}
	\vspace{-2ex}
\end{figure}

% Introduction of roles and tasks
Teams participating in the TeenSize class in RoboCup 2017 can be composed by a maximum of three robots: one goalkeeper and two field players. 
We define dynamic \textit{Player Tasks} which are frequently reassigned during the game.
This task tells the robot what it is supposed to do according to its own state in the field and the state of its teammates.
We define the following tasks: Attack, Defend, KeepGoal, ChangeTask and WaitClearOut.
In addition, we define a task manager which is in charge of the safe assignment of these tasks.
Each of this tasks is associated with a respective state of the game FSM (Fig. \ref{fig:roles_tasks}). 

\subsubsection{Attack Task}
A robot with this task, known as \textit{striker}, has active interaction with the ball. 
In possession of the ball, the robot will try to score either by kicking directly or by dribbling to get a better position for kicking the ball.
The robot will also reach the ball and search for the ball in case the robot does not possess it.
Searching for the ball, the robot goes first to the place where the ball was last seen. 
It turns on the spot and if the ball location is still unknown, it will go to the penalty marks, first to the closest and then to the furthest one.
Reaching the ball means to place the robot behind the ball so it can kick or dribble.
When approaching the ball, the robot does consider the ball as an obstacle in order to avoid undesired hits.

\subsubsection{Defend Task}
\begin{figure}[t]
	\centering
	\includegraphics[width=0.9\linewidth]{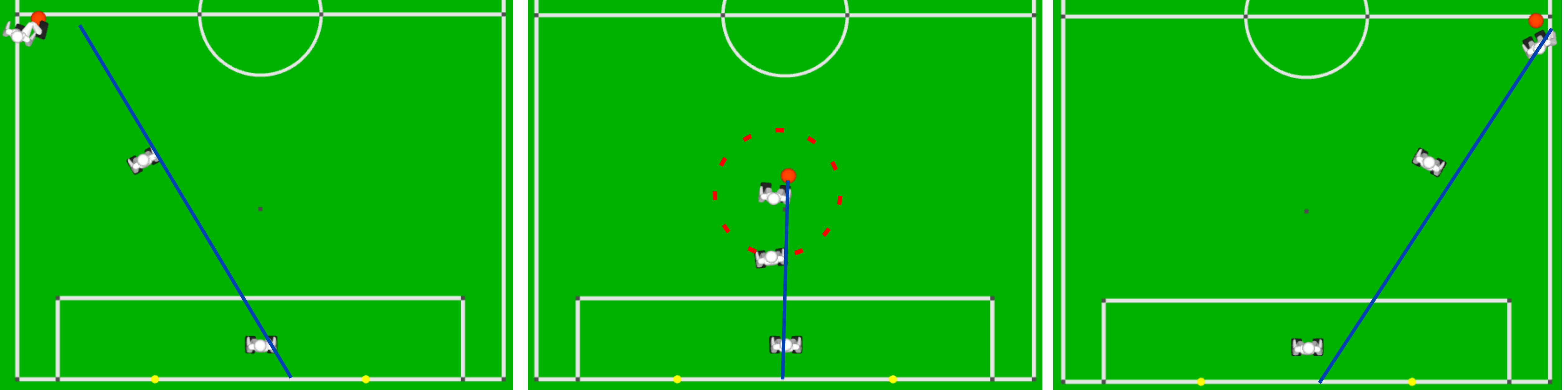} 
	\caption{Different poses that the defender might have according to the ball pose. On the middle the minimum distance between field players is shown in red.}
	\label{fig:defender}
	\vspace{-2ex}
\end{figure}

\begin{figure}
	\centering
	\includegraphics[width=1.0\linewidth]{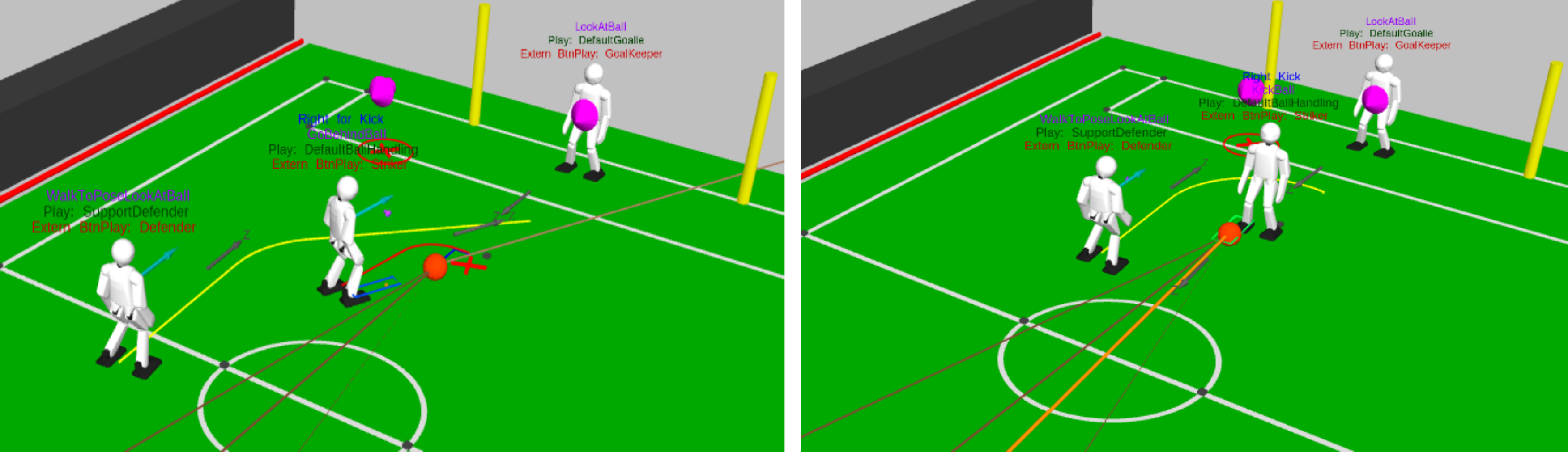} 
	\caption{Collision avoidance between field players. The defender considers the pose of the striker and surrounds it.}
	\label{fig:defender_features}
	\vspace{-4ex}
\end{figure}

The \textit{defender} robot is not supposed to have contact with the ball but to be ready to change its task and approach the ball if necessary.
Its position is defined by a vector coming from the middle of its own goal towards the position of the ball.
The magnitude of this vector is defined proportional to the distance of the ball to the own goal and saturated in order to avoid collisions with team members.
In case the ball is not visible, this magnitude is calculated using the pose of the striker.
In this manner, the robot is able: i) to block opposite direct shots, ii) to be ready for one-vs-one fights, and iii) to get possession of the ball in case the previous striker is taken out of the match.
With respect to the orientation, the robot tries to look in the direction of the ball.
Figure \ref{fig:defender} shows the pose of the defender for different ball locations.
In order to avoid collisions with other teammates, the initial shortest path (straight line) is modified such that any teammate in between the path is surrounded (Fig. \ref{fig:defender_features}).

\subsubsection{KeepGoal Task}
The behavior of this task mainly depends on the proximity of the ball, of the opponents, and of the teammates to the own goal.
Because the other teammates have to coordinate their behaviors according to what the goalkeeper is doing, the decision if the goalkeeper has to clear out the ball is managed by the task manager.
When the goalkeeper receives the signal to clear out the ball, it goes to the ball and hits it in direction of the opposite goal.
It can also get a signal to dive including the diving direction.
On the other hand, if the task manager decides that the goalkeeper does not need to clear out the ball, the robot will only move laterally on the same horizontal level of the ball position.

\subsubsection{WaitClearOut}
Different to the FIFA rules, the RoboCup Humanoid League has a special rule that prohibits more than one robot to be in the own area for more than \unit[10]{s}. 
For this reason, when the ball is the own goal area, an additional coordination between team members is required.
When a robot announces that it will clear out the ball from the own goal area, a \textit{WaitClearOut} task reassignment occurs. 
In this task, the robots go closer to the ball without entering the own goal area to avoid the illegal defense.
The path is planned such that the robot will not block the shot from an other robot clearing out the ball.
In addition, the target pose in this task is assigned such that two robots will not collide with each other.
\begin{figure}[b]
	\centering
	\includegraphics[width=1.0\linewidth]{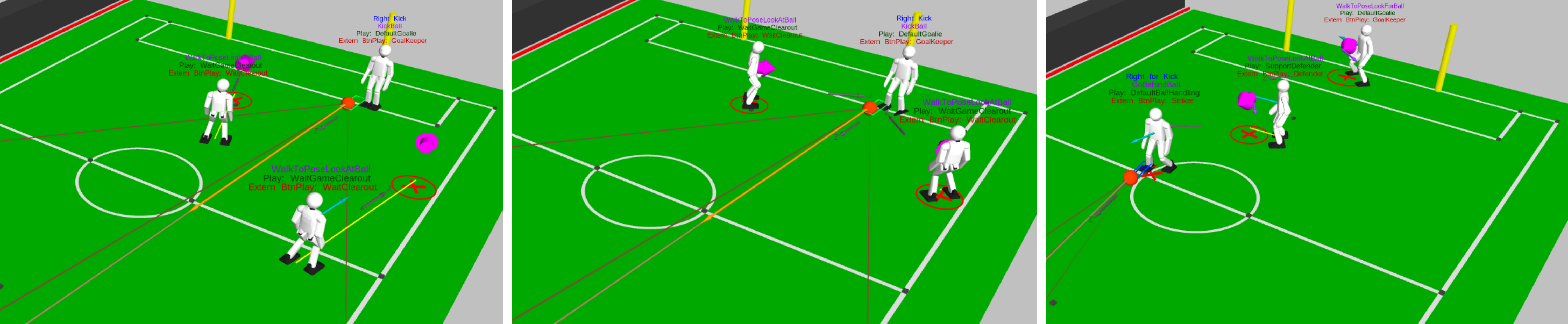} 
	\caption{Ball clear out by the goalkeeper. The yellow line represents the planned path, while the surrounded red cross refers to the instantaneous 2D target pose. Note that the robots always look at the ball. The figure at the rightmost shows the game after the clear out.}
	\label{fig:wait_clearout}
	\vspace{-2ex}
\end{figure}

% Task Mananger
The task manager of each robot determines the desired task and triggers special events that have to be coordinated.
The desired task depends on each role. 
A goalkeeper only has assigned the task \textit{KeepGoal}, while a field player can alternate between \textit{Attack}, \textit{Defend}, \textit{ChangeTask} and \textit{WaitClearOut}.
In order to handle possible noise or very fast alternating data that could lead to a continuous task change, the request is only made once the decision is confirmed for a defined number of consecutive cycles.
This voting system is persistent for all decisions taken by the task manager.
If there is only one field player in the match, it will be assigned to \textit{Attack}.
The robot with the task of \textit{Defend} can request a task change to the striker, but not vice versa.
If the defender estimates that it is the closest to the ball, it will request a task change.
When a striker leaves the field, because of a service or a penalization, it announces its egress such that a task assignment can be done immediately.

\begin{wrapfigure}{R}{0.5\textwidth}
\vspace*{-2ex}\includegraphics[width=\linewidth]{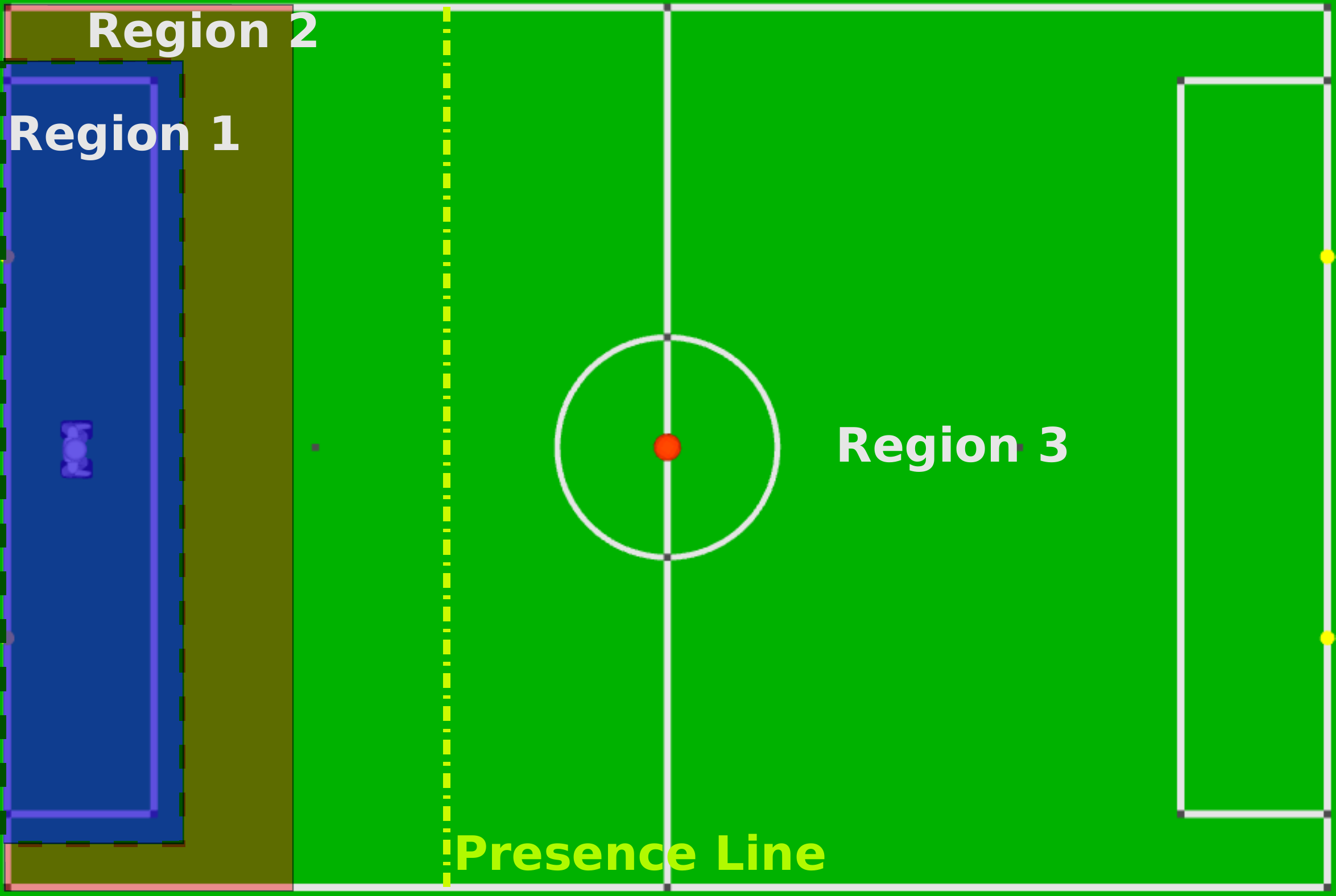} 
	\caption{Areas determining  goalkeeper ball clearing behavior. In Region~1, goalkeeper needs to clear out the ball. In Region~3, the robot is remains in its goal. In Region~2, the goalkeeper clears out the ball if there is no field player between the own goal and the presence line (yellow dotted line).}
	\label{fig:goalie_clearout}
	\vspace{-2ex}
\end{wrapfigure}

% Task manager of the clearing out action for the goalie
Clearing out the ball by the goalie is a special event to be handled by the task manager because it implies task changes of the field players.
In this case, the rest of the field players change to the \textit{WaitClearOut} task, that place them in a strategic position once the ball has been cleared out.
The decision of clearing out the ball by the goalie depends mostly on the ball location.
The field is divided in three regions (Fig. \ref{fig:goalie_clearout}).
The first region is the goal area with an additional outer tolerance.
In goal area, the goalie gets the highest priority, and when it is asked to clear out the ball, the field players set their task to \textit{WaitClearOut}.
The second region limits the possible areas where the goalkeeper can clear out the ball.
If the ball is further away, the goalie just waits for its teammates.
In this region, however, the goalie only chases the ball to clear it out if there is no teammate close to the own goal, i.e., if there is no teammate between our goal and the presence line (Fig. \ref{fig:goalie_clearout}).
In the third region, the goalkeeper takes a more passive role and gazes at the ball to be ready for diving.

% State and Task Assignment
The task assignment is based on an asynchronous request-and-response system that ensures that there is only one robot actively interacting with the ball.
This prohibits, for example, that two robots try to kick the ball simultaneously which could lead to team self-collisions.
The request for a task reassignment depends on the state of the robot and its teammates.
This state comprises current task, ball distance, ball visibility, ball possession, ball location, current robot pose, if the robot is active and if the robot is not fallen.
This information is estimated using the team communication data broadcast at a rate of \unit[8]{Hz}.
Only recent data (received in the last \unit[5]{seconds}) is, however, considered because of possible hardware failures or communication errors.
\begin{figure}[t]
	\centering
	\includegraphics[width=1.0\linewidth]{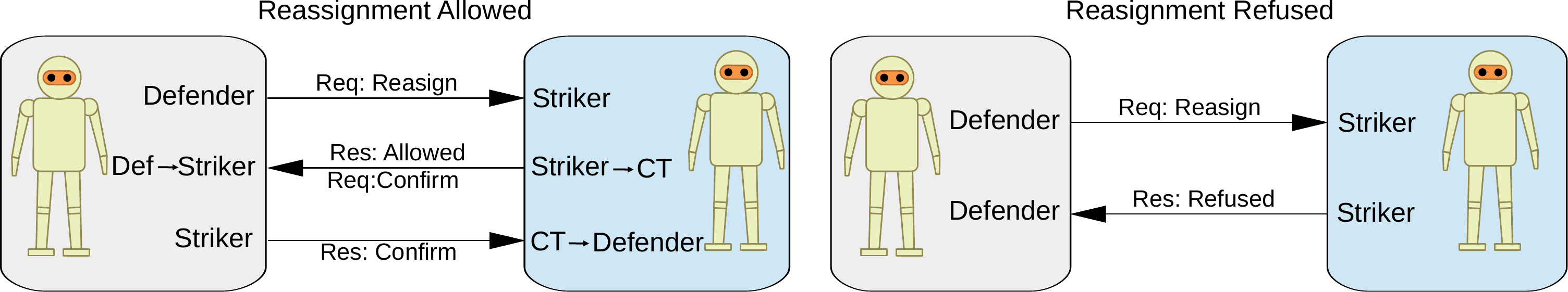} 
	\caption{Task assignment. A defender sends a request to change task. If the request is accepted, the striker is assigned to ChangeTask. It requests a confirmation, and the defender changes its task. If the request is refused, both robots keep their current tasks.}
	\label{fig:task_assignment}
	\vspace{-2ex}
\end{figure}

% Realization of task assignment
For task reassignment, the robot with the \textit{Attack} task has a higher priority over \textit{Defend}, i.e., a defending robot can only change its task if it is allowed by the current striker.
When the defender finds itself in a better position to possess the ball, it requests to change tasks.
If the striker also determines that its teammate is in a more convenient position, the striker changes its task to \textit{ChangeTask} and sends back a response. 
The requester changes correspondingly its task and sends a confirmation such that the robot with task \textit{ChangeTask} can change to the new task.
On the other hand, if the original striker finds that it is in a better position to possess the ball, it sends back a negative response and no task reassignment takes place (Fig. \ref{fig:task_assignment}). 

\begin{figure}[b]
	\centering
	\includegraphics[width=0.5\linewidth]{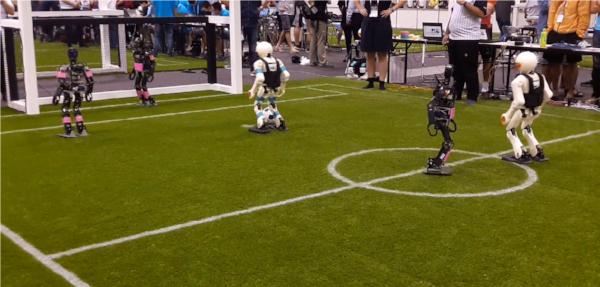} 
	\includegraphics[width=0.47\linewidth]{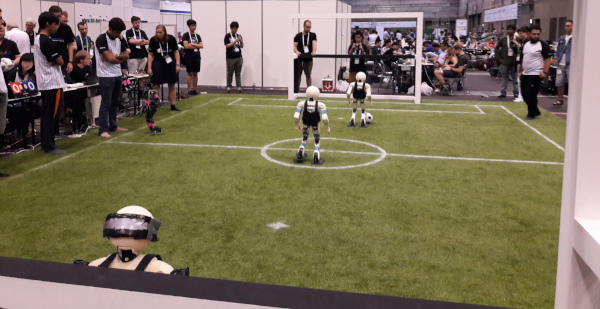}\vspace*{-1ex} 
	\caption{Team play in RoboCup 2017.}
	\label{fig:robocup_teamplay}
	\vspace{-2ex}
\end{figure}

The team play strategies presented here were used in the RoboCup2017 competition.
Figure \ref{fig:robocup_teamplay} shows official matches in which the striker and defender roles can be distinguished.

\subsection{Landing Motions}
\begin{figure}
\parbox{\linewidth}{
\centering
\includegraphics[width=0.24\linewidth]{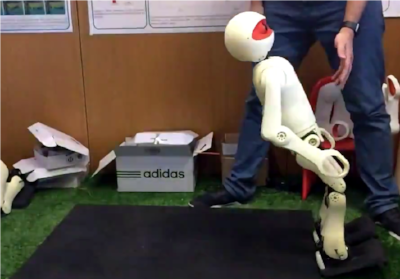}
\includegraphics[width=0.24\linewidth]{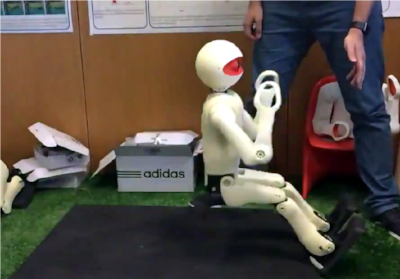}
\includegraphics[width=0.24\linewidth]{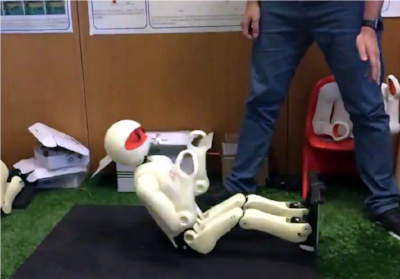}
\includegraphics[width=0.24\linewidth]{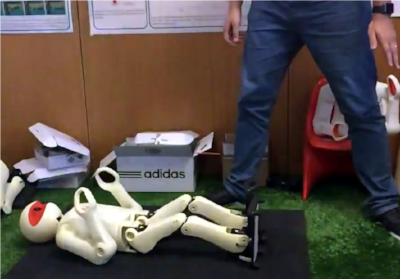}\vspace{2pt}
\includegraphics[width=0.24\linewidth]{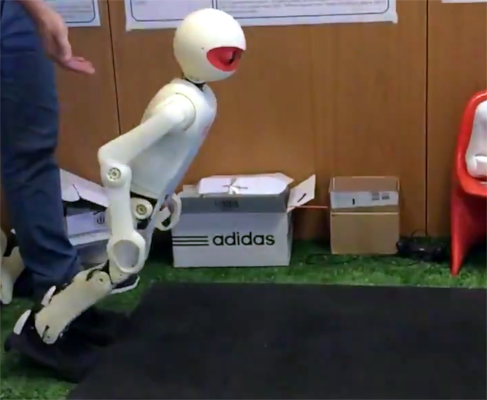} 
\includegraphics[width=0.24\linewidth]{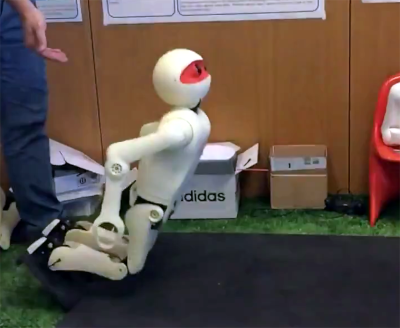}
\includegraphics[width=0.24\linewidth]{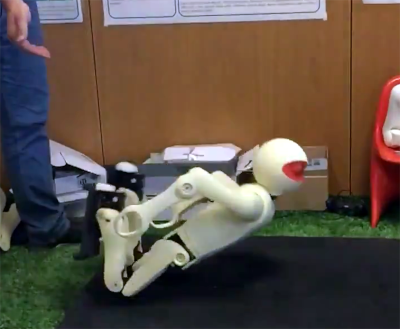}
\includegraphics[width=0.24\linewidth]{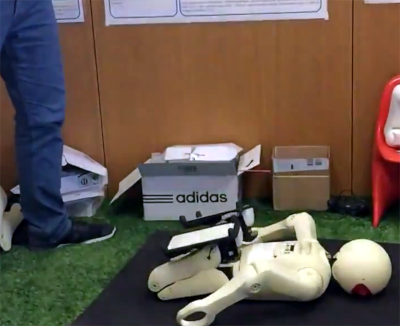}
}
\caption{Landing Motions in forward and backward direction.}
\label{fig:landing_motion}
\vspace{-2ex}
\end{figure}

In a match during the competition, the robot might fall even with a robust and fine-tuned gait. 
The actions of our opponent can lead to difficult situations as happened in RoboCup 2016 \cite{Farazi2017}.
This implies that the robot needs to know how to land.
We designed landing motions that are activated when the robot would face an inevitable fall situation either in forward or backward direction.
The aim of these motions is to protect the hardware of the robot from impacts in delicate parts of the robot, e.g. the knees.
The designed motions are shown in Fig. \ref{fig:landing_motion}.
%-----------------------------------
%-----------------------------------

%-------------------------------------------------------------------------------
\subsection{Human-Robot Interfaces}

To configure and calibrate the robots, a web application system is used with the robot PC as the web server.
Even over a poor quality wireless network, the connection is robust by making use of the client-server architecture of web applications and the highly developed underlying web protocols.
In addition, most of the processing is carried out by the client, resulting in a low computational cost for the robot.
The web application allows the user: to start, to stop, to monitor ROS nodes, to display status information, to dynamically reconfigure the robot, and to visualize the processed vision, amongst others.
%%%%%%%%%%%%%%%%%%%%%%%%%%%%%%%%%%%%%%%%%%%%
\section{Game Performance}
In RoboCup 2017, our robots scored 37 goals in 12 matches and did not receive any goal.
27 goals were scored during seven games of the TeenSize tournament and the remaining 10 goals were scored in five Drop-in games.
This proved the robustness of the methods presented in this paper.
Our robots won all the matches in the TeenSize tournament including the final 2:0 vs. HuroEvolutionTN (Taiwan).
In the Drop-in games, the individual skills were tested and our robots have obtained 21 points in total, having a margin of 19 points to the team in second place.

%%%%%%%%%%%%%%%%%%%%%%%%%%%%%%%%%%%%%%%%%%%%
\section{Technical Challenges}
\seclabel{technical_challenges}

%-----------------------------------
\begin{figure}[b]
\parbox{\linewidth}{
\centering\footnotesize
a)\,\includegraphics[width=0.25\linewidth]{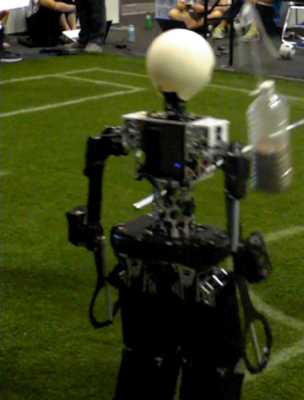}
\hspace{0.019\linewidth}
b)\,\includegraphics[width=0.25\linewidth]{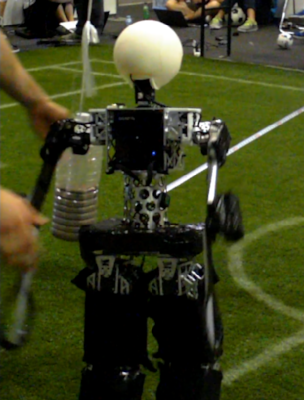}
\hspace{0.019\linewidth}
c)\,\includegraphics[width=0.25\linewidth]{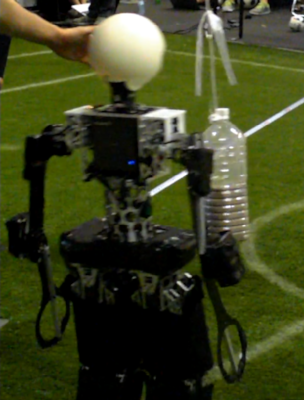}\vspace*{1ex}
}
\caption{Dynaped withstanding a push from the front. 
a) Before impact. 
b) Immediately after impact. One can see that the robot is unstable.
c) Stable posture is recovered.}
\figlabel{push_recovery}
\vspace{-2ex}
\end{figure}
%-----------------------------------
In RoboCup 2017 there were four technical challenges.
They were designed to evaluate specific capabilities of robots in isolation, separately from the regular games.
In this section we briefly describe our strategies for these challenges.
%===============================================================================
\vspace{-2ex}
\subsubsection{Push Recovery}
\seclabel{push_recovery}
In this challenge, the robot is pushed from the front and from the back. 
The goal is to withstand the impact and recover a stable posture.
In RoboCup 2017, Dynaped successfully completed this challenge, withstanding the push from a 1.5\,kg pendulum which was retracted by 55\,cm (\figref{push_recovery}).
%==============================================================================
\vspace{-2ex}
\subsubsection{High Jump}
\seclabel{high_jump}
In this challenge, the task is to jump as high as possible and remain in the air as long as possible.
In order to perform this task, a jumping motion was designed using our keyframe editor.
In RoboCup 2017, one of ours \iguhop robot successfully performed a jump of height 4.5\,cm, remaining 0.192\,s in the air and stand stable afterwards.
%===============================================================================
\vspace{-2ex}
\subsubsection{High Kick}
\seclabel{high_kick}
The goal of this challenge is to kick the ball over the obstacle into the goal.
In order to complete this challenge and overcome as high an obstacle as possible, a modified foot was used by one of our \iguhop robots for kicking.
The foot had a smooth concave shape which allowed it to scoop the ball effectively and, hence, kick it upwards, overcoming the obstacle.
%The kicking motion is played using keyframes and was predefined with the keyframe editor.
In RoboCup 2017, our robot was able to complete this challenge with a height of \unit[8]{cm}.
The execution of a high kick over a 21\,cm obstacle, recorded during testing in our lab, is shown in \figref{high_kick}.
%-----------------------------------
\begin{figure}[t]
\parbox{\linewidth}{\centering\footnotesize
a)\,\includegraphics[width=0.3\linewidth]{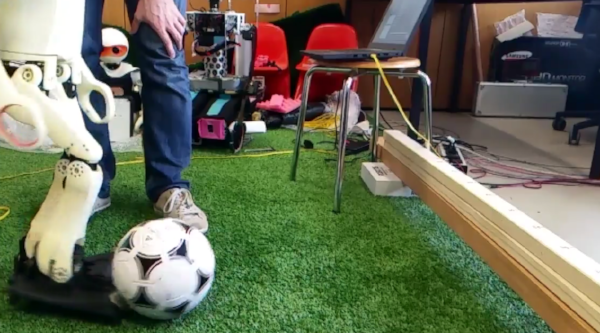}~b)\,\includegraphics[width=0.3\linewidth]{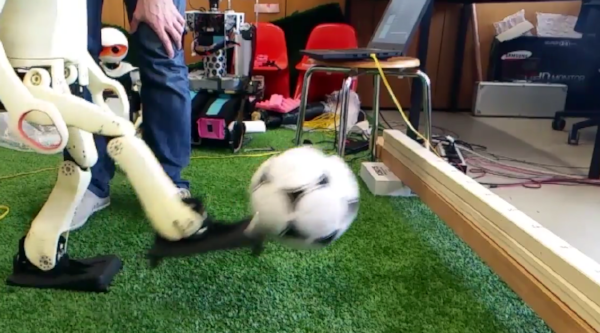}~
c)\,\includegraphics[width=0.3\linewidth]{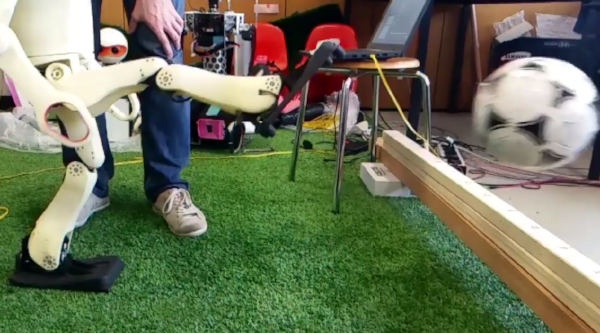}}\vspace*{-1ex}
\caption{Robot performing a high kick over a 21\,cm obstacle.
a) The ball is being scooped by the foot.
b) The ball is kicked upwards.
c) The ball overcame the obstacle.}
\figlabel{high_kick}
\vspace{-2ex}
\end{figure}
%-----------------------------------
\begin{figure}[b]
\parbox{\linewidth}{\centering
\includegraphics[width=0.3\linewidth]{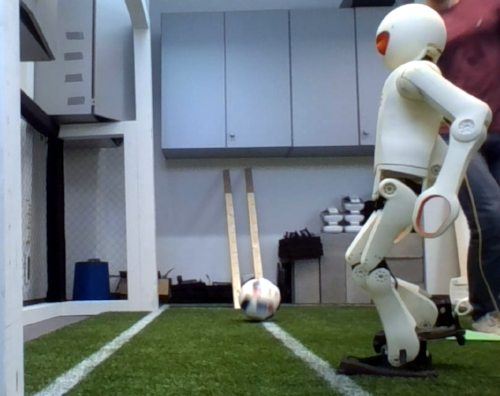}
\hspace{0.019\linewidth}
\includegraphics[width=0.3\linewidth]{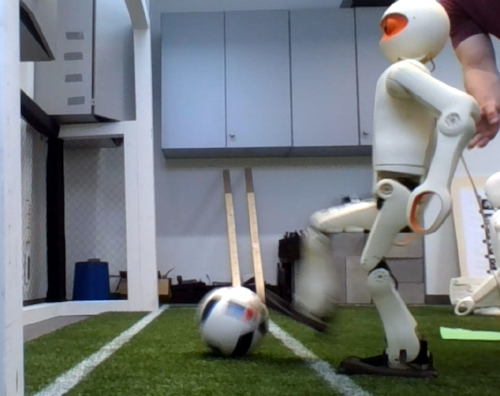}
\hspace{0.019\linewidth}
\includegraphics[width=0.3\linewidth]{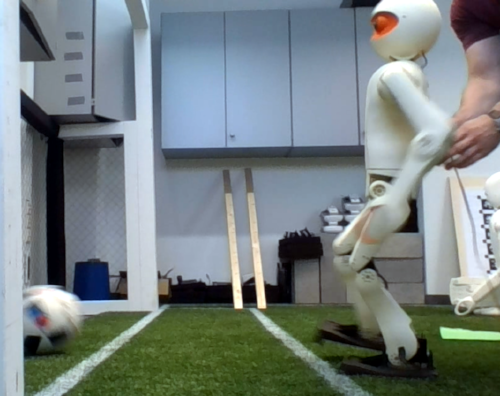}}
\caption{Robot scoring from a moving ball.
Left: The robot is waiting for the ball.
Middle: The kick is executed.
Right: A goal is scored.}
\figlabel{moving_ball}
\vspace{-2ex}
\end{figure}
\vspace{-2ex}
\subsubsection{Goal Kick from Moving Ball}
\seclabel{moving_ball}
The task of this challenge is to score a goal by kicking the moving ball into the goal.
The ball is rolling along the goal area line.
We solved this task as follows: first, we shift all weight onto one of the legs while having the other leg lifted and ready to kick; then, we estimate the velocity of the rolling ball and hence, time of arrival thereof to the kicking region; and, finally, we perform the kick according to the previous estimate.
Following this strategy our robot was able to successfully complete this challenge.
Our robot, performing this task during tests in our lab, is shown in \figref{moving_ball}.

%%%%%%%%%%%%%%%%%%%%%%%%%%%%%%%%%%%%%%%%%%%%
\section{Conclusions}
In this paper, we presented the methods and approaches that lead us to win all possible competitions in the TeenSize class for the RoboCup 2017 Humanoids League in Nagoya: the soccer tournament, the Drop-in games, and the technical challenges.
We presented individual skills regarding the perception and the bipedal gait, and their application in the technical challenges.
Additionally, team skills were also extensively explained.

\subsection*{Acknowledgements}\vspace*{-1ex}
\footnotesize
This work was partially funded by grant BE 2556/13 of German Research Foundation.

%%%%%%%%%%%%%%%%%%%%%%%%%%%%%%%%%%%%%%%%%%%%
\bibliographystyle{IEEEtran}
\bibliography{winners_2017}

%%%%%%%%%%%%%%%%%%%%%%%%%%%%%%%%%%%%%%%%%%%%
\end{document}